\documentclass[10pt,twocolumn,letterpaper]{article}

\usepackage{iccv}
\usepackage{times}
\usepackage{epsfig}
\usepackage{graphicx}
\usepackage{amsmath}
\usepackage{amssymb}

\usepackage{array, booktabs}
\usepackage{comment}
\usepackage{algorithm}
\usepackage{algorithmic}
\usepackage{bbding}
\usepackage{pifont}
\usepackage{wasysym}
\usepackage{amssymb}
\usepackage{braket}

\usepackage{xcolor}

\usepackage[pagebackref=true,breaklinks=true,letterpaper=true,colorlinks,bookmarks=false]{hyperref}

\iccvfinalcopy 

\def\iccvPaperID{12256} 

\ificcvfinal\fi

\begin{document}

\title{RefEgo: Referring Expression Comprehension Dataset \\ from First-Person Perception of Ego4D}

\author{Shuhei Kurita$^{1}$\thanks{Corresponding author.}  ~~~~~~~~~~~~~~~~~~~~~~~  Naoki Katsura$^{2,1}$ ~~~~~~~~~~~~~~~~~~~~~~~ Eri Onami$^{3,1}$\\
$^{1}$RIKEN  ~~~~~  $^{2}$University of Tsukuba ~~~~~ $^{3}$Nara Institute of Science and Technology\hspace{-6em}\\
{\tt\small shuhei.kurita@riken.jp, ~ n-a-katsura@mercari.com, ~ onami.eri.ob6@is.naist.jp}
}


\maketitle
\ificcvfinal\thispagestyle{empty}\fi

\begin{abstract}

Grounding textual expressions on scene objects from first-person views is a truly demanding capability in developing agents that are aware of their surroundings and behave following intuitive text instructions. Such capability is of necessity for glass-devices or autonomous robots to localize referred objects in the real-world. In the conventional referring expression comprehension tasks of images, however, datasets are mostly constructed based on the web-crawled data and don't reflect diverse real-world structures on the task of grounding textual expressions in diverse objects in the real world. Recently, a massive-scale egocentric video dataset of Ego4D was proposed. Ego4D covers around the world diverse real-world scenes including numerous indoor and outdoor situations such as shopping, cooking, walking, talking, manufacturing, etc. Based on egocentric videos of Ego4D, we constructed a broad coverage of the video-based referring expression comprehension dataset: RefEgo. Our dataset includes more than 12k video clips and 41 hours for video-based referring expression comprehension annotation.
In experiments, we combine the state-of-the-art 2D referring expression comprehension models with the object tracking algorithm, achieving the video-wise referred object tracking even in difficult conditions: the referred object becomes out-of-frame in the middle of the video or multiple similar objects are presented in the video. Codes are available at \url{https://github.com/shuheikurita/RefEgo}

\end{abstract}



\section{Introduction}
\label{sec:intro}

It is a truly demanding task to identify surrounding objects in real world scenes from video clips of egocentric viewpoints with free-form language supervisions. Such a task is necessary for glass-devices or autonomous robots that help with daily-life tasks and communicate with us in language because they need to understand the intuitive expressions of languages and ground them into the surrounding world. It is an ultimate goal for the referring expression comprehension (REC) or shortly ``visual grounding'' task because it maps the referred entities in text to the corresponding objects identified and tracked from the observed sequence of images.

Extensive efforts are being made in 2D image reference expression comprehension~\cite{referitgame, flickr30k,refcoco,refcocog}. Recent semi-supervised approaches contribute to the open-vocabulary object detection from 2D images~\cite{mdetr,detic}. However, compared to these extensive studies on 2D image referring expression comprehension, we notice that comparably less efforts are taken in video-based referring expression comprehension~\cite{lingualotb99,Khoreva2018VideoOS,referYouTube-VOS}. Video clips in such datasets are mostly collected in the Internet and aren't suitable for the real-world daily-task understandings. The number of video clips are also limited.
Ideally, video clips for such tasks are collected in embedded form in our daily lives and cover variety domains such as walking streets, shopping,  chatting with others, staying in indoor, cleaning laundries, or cooking foods,
when we pursue general purpose models in our daily scenes.
However, it was nearly prohibitive to create datasets on such tasks because of the lack of the collection of real-world setting egocentric videos.

\begin{figure}[t]
    \centering
    \includegraphics[width=8cm]{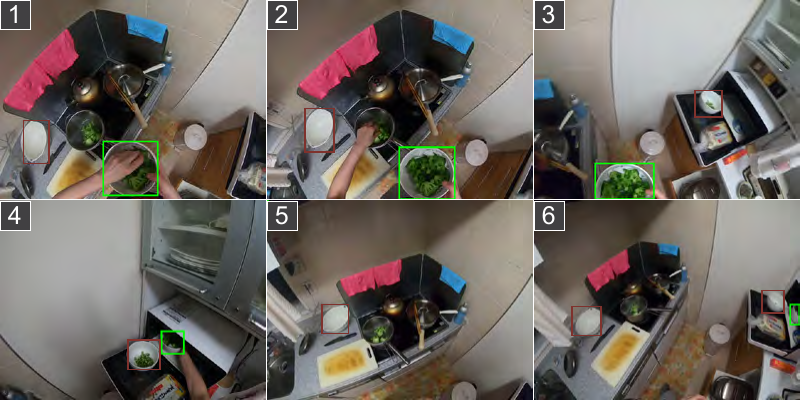}
    \caption{Sample frames of RefEgo for ``the large white bowl with broccoli inside that is used to load the pan of broccoli.'' The referred object of the bowl in green and other bowls in brown. }
    \label{fig:teaser}
\end{figure}

\renewcommand{\arraystretch}{0.93}
\begin{table*}[t]
\begin{center}
\footnotesize\begin{tabular}{lcccccc}
    \toprule
Video REC   & Base Dataset    &\# Clips  & \# Object Annotations & \# Objects & \#  Categories   \\
\midrule
Lingual OTB99~\cite{lingualotb99}           & OTB100~\cite{otb100}       &  99  & 58,733 & 99  & - \\
Lingual ImageNet Videos~\cite{lingualotb99} & ImageNet VID~\cite{imagenetvid} & 100  & 23,855 & 100 & 25 \\
\midrule
Video Object Segmentation & Base Dataset  &\# Clips & \# Object Annotations &  \# Objects & \# Categories  \\
\midrule
ReferDAVIS-16~\cite{Khoreva2018VideoOS}     & DAVIS~\cite{davis}  & 50 & 3,440 & 50  & - \\
ReferDAVIS-17~\cite{Khoreva2018VideoOS}     & DAVIS~\cite{davis}  & 90 & 13,540 & 205 & - \\
Refer-Youtube-VOS~\cite{referYouTube-VOS}   & Youtube-VOS~\cite{YouTube-VOS}    & 3,252 & 133,886 & 6,048 & 78 \\
\midrule
RefEgo (ours)& Ego4D~\cite{Ego4D2022CVPR} (First-person video) & 12,038 & 226,319 & 12,038  & 505 \\
\bottomrule
\end{tabular}
    \caption{
        Comparison of the video-based referring expression comprehension datasets.
    }
    \label{table:video_rec_datasets}
\end{center}
\end{table*}

\renewcommand{\arraystretch}{0.93}
\begin{table}[t]
\begin{center}
\footnotesize\begin{tabular}{lcc}
\toprule
REC dataset    & \# Images  &\# Object Annotations \\
\midrule
RefCOCO~\cite{refcoco}  & 19,994 & 50,000  \\
RefCOCO+~\cite{refcoco} & 19,992 & 49,856  \\
RefCOCOg~\cite{refcocog}& 26,711 & 54,822  \\
\midrule
RefEgo (ours) & 226,319 & 226,319  \\
\bottomrule
\end{tabular}
    \caption{
        Comparison of the RefEgo dataset to 2D REC datasets. RefCOCO/+/g~ datasets are based on MSCOCO~\cite{mscoco} images while ours is based on real-world egocentric video of Ego4D.
    }
    \label{table:2d_rec_datasets}
\end{center}
\end{table}

Recently, Ego4D~\cite{Ego4D2022CVPR}, a massive-scale collection of egocentric video and annotation is proposed.
Ego4D videos are gathered by 931 unique participants in 74 locations worldwide.
They are captured by a head-mounted camera device, intending to capture various daily-life activities in first person vision, covering hundreds of daily life activities including various locations: in-store, office-space, houses, wire-house, street and so on.
Based on Ego4D videos, we constructed the novel in a margin larger RefEgo video-based referring expression comprehension dataset with the help of object detection and human annotation, aiming to ground intuitive language expressions on various contexts in real-world first person perception.

Our RefEgo dataset exhibits unique characteristics that make it challenging to localize the textually referred objects. It is based on egocentric videos and hence includes frequent motions in video clips. The referred objects are often surrounded by other similar objects of the same class. The referred object may appear at the edge of the image frame or even goes out-of-frame in some frames, requiring models to discriminate images that contain and don't contain the referred object.
Fig.~\ref{fig:teaser} presents the selective frames from a single video clip with a referred expression of ``the large white bowl with broccoli inside that is used to load the pan of broccoli.''
There are several other bowls in image frames and the referred object goes out-of-frame in the fifth frame.
In this case, the models are expected to predict that the referred object is not presented in the image, illustrating the challenging characteristics of the proposed RefEgo.

We prepared valuable baseline models from multiple approaches for RefEgo.
We applied the state-of-the-art REC models of MDETR~\cite{mdetr} and OFA~\cite{ofa} for RefEgo. We introduced MDETR models trained with all images including image frames with no annotated referred objects, and observed it performs better at discriminating images without referred objects than other models.
We also introduce MDETR with a special binary head to discriminate images without the referred object presented.
We finally apply the object tracking of ByteTrack~\cite{bytetrack} for combining multiple detection results, allowing the models to spatio-temporal localization of the referred object.

\section{Related Work}
\label{sec:related_work}

\subsection{Referring expression comprehension}

Grounding textual expressions in objects is a key component in vision-and-language studies.
It includes several formalism depending on the data type of visual information reflecting the spatial and temporal diversity of the real-world data: a single image, sequence of images or video clips, and 3D reconstructed data.
Referring expression comprehension in image, or simply \textit{visual grounding}, is a task to localize objects in an image from open vocabulary texts~\cite{referitgame, flickr30k,refcoco,refcocog,Liu_2019_CVPR, phrasecut}.
This is one of the most active research field in the vision and language field and many advanced approaches are proposed in these years~\cite{yu2018mattnet,muchen2021referring,mdetr,ofa}.
However, as these studies are limited to a given single image and hence these REC models have limited knowledge of the referred object and its surrounding environments.
3D-scene based approaches for referring expression are also another major branch in real-worlds scene grounding~\cite{ScanRefer,Wald2019RIO,reverie,episodic_mem_qa,azuma_2022_CVPR}.
Although there are numerous benefits in 2D and 3D referring expression comprehension datasets, these spatial datasets lack of the temporal localization in the real-world.

\subsection{Video-based REC}

\paragraph{Language-based object tracking}
Video based localization is a both temporal and spatial localization of objects in video frames.
These datasets are often provided as a language-annotation extension to the existing dataset and changes the core concept to determine what object to track.
Lingual OTB99 and ImageNet Videos~\cite{lingualotb99} are language-based object tracking datasets that are based on existing object tracking dataset~\cite{otb100,imagenetvid}. Here language annotation is attached for first image to specify the object to track in later frames.
For limited domain sets, person category annotation is performed~\cite{Yamaguchi2017SpatioTemporalPR}.
VID-Sentence dataset~\cite{VID-Sentence} also annotated in a part of the ImageNet VID dataset.
Co-grounding network~\cite{cogrounding_2021_CVPR} and DCNet~\cite{cao2022correspondence} is proposed for these video-based REC dataset.
It is also notable that while multi-object tracking datasets~\cite{MOTChallenge2015} tend to cover limited object classes, the object tracking dataset of TAO~\cite{tao2020} includes 894 objects of 345 free-form text classes in a part of their dataset. 

\begin{figure*}[t]
    \centering
    \includegraphics[width=15cm]{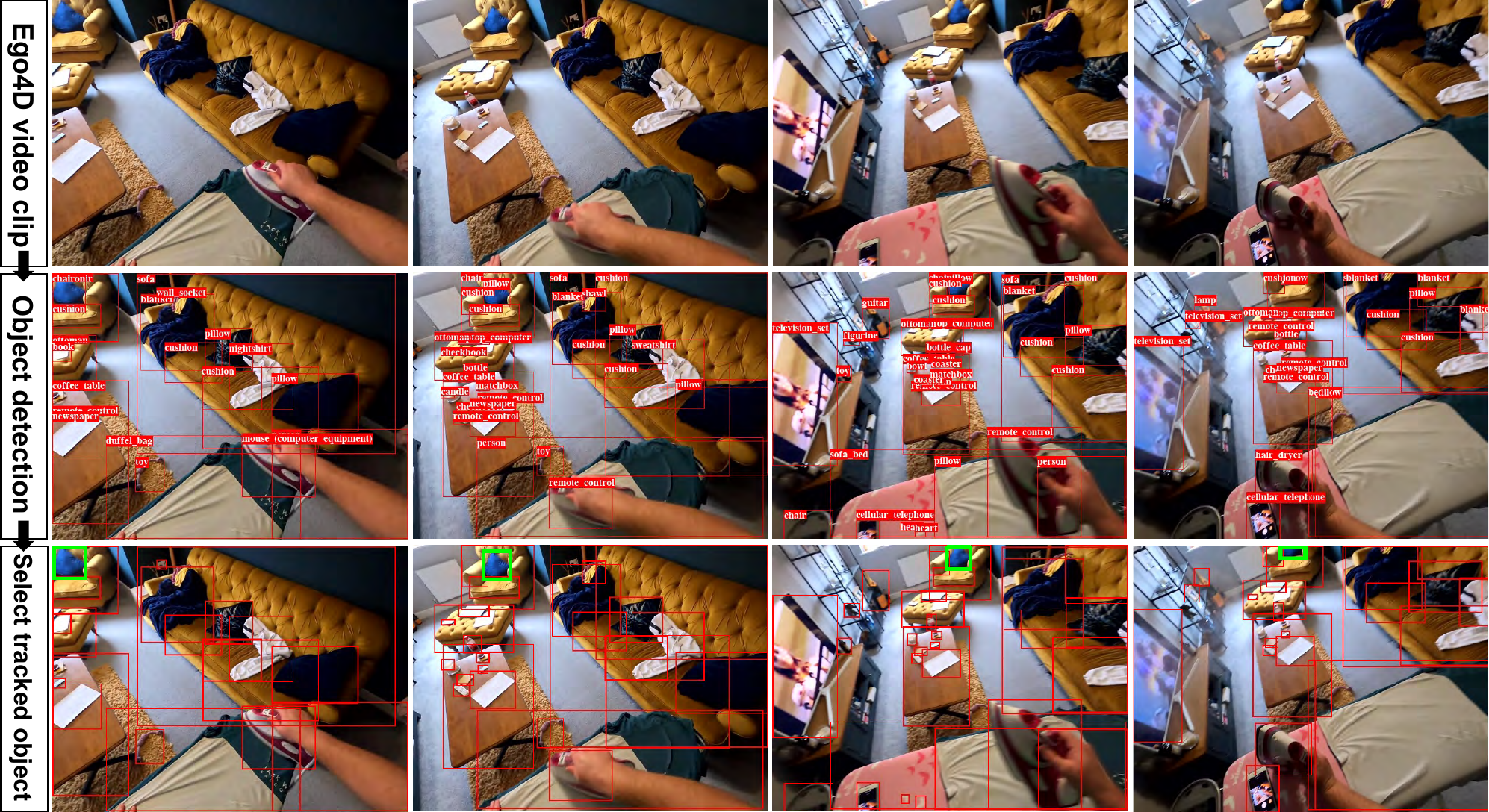}
    \caption{The process of attaching the bounding box for the blue pillow placed far distant in a Ego4D video clip.
    The annotated referring expression is ``a square bright blue pillow on the chair in front of the ottoman.'' }
    \label{fig:Ref-Ego4d wordcloud}
\end{figure*}

\paragraph{Language-based video object segmentation}
Video object segmentation (VOS) is the segmentation task of a target object in a video clip~\cite{Khoreva2018VideoOS,referYouTube-VOS}. These tasks are constructed on the existing VOS datasets~\cite{davis,YouTube-VOS}.
In conventional setting, the target object is specified by a pixel-accurate mask, while the language-based localization annotation was proposed~\cite{Khoreva2018VideoOS,referYouTube-VOS} for language-based specification instead of the pixel masks.
\\

Our dataset is based on first-person videos and have specific characteristics discussed in Sec.~\ref{sec:task}.
We provide further detailed comparisons with language-based object tracking and video segmentation datasets in Table~\ref{table:video_rec_datasets} and the scale comparisons with the conventional 2D REC datasets in Table~\ref{table:2d_rec_datasets}.
Our dataset is in a magnitude larger than the existing REC datasets in terms of the number of the total object annotations and the unique objects tracked.

\subsection{Ego4D episodic memory benchmark}

The Ego4D dataset provides the episodic memory benchmark that aims to query videos and localize the answer of the query.
They provide the \textit{visual query}-based VQ2D and VQ3D tasks, the \textit{natural language queries}-based NLQ task for determining the temporal window of the video history where the query answer is evident, and the \textit{moments queries} of MQ task for localizing all instances of the given activity name.
Among these tasks, the NLQ task is based on the flexible natural language query.
However, there are steep differences between NLQ and our task: in the NLQ task, models localize the temporal window of the occurrence (e.g., ``What did I put in the drawer?''), while in our dataset, models localize the directly referred object (e.g.,  ``the cushion on the right end of the sofa'') in spatial and temporal manner. As it requires explicit spatial localization and tracking for all temporal frames where the referred object is presented, the video clips of our dataset become shorter than NLQ.
Our focus is on creating a comprehensive dataset for first-person video-based referring expression comprehension, which aims to not only concentrate on the different aspects from the Ego4D episodic memory benchmarks, but also contribute them by providing rich annotated data for natural language-based object localization and tracking.



\section{RefEgo Dataset}
\label{sec:dataset}

\subsection{Task}
\label{sec:task}

Our task is to localize a textually referred object in a sequence of images from an egocentric video.
Given a single phrase of a referring expression for one target object referred, the model predicts bounding boxes for the referred object in a sequence of images drawn from the Ego4D videos.
As the first-person videos include high viewpoint motions, the referred objects sometimes locate out of frames for some images.
Therefore, we introduce the task of \textit{discriminating images that do not include the referred object} from other images that include the referred object when the models are given a sequence of images. This is in addition to the conventional object localization task by \textit{predicting a single bounding box of the referred object for images that include the referred object}.
We summarize the special conditions that make our RefEgo dataset plausible in real-world experiments as follows.

\noindent
\textbf{Frequent viewpoint motion}
Ego4D videos are captured by wearable cameras and hence experience a number of viewpoint motions.
The frequent viewpoint motion makes it challenging to identify the same objects in distinct frames.
Here, we assume the REC models can help the object tracking methods as the tracking-by-detection approach.
If REC models are accurate enough, it can successfully ground the single target object in images onto the unique referred expressions, therefore they simultaneously solve the visual grounding and object tracking problems, independent of the frequent viewpoint motion.

\noindent
\textbf{Detection of the no-referred-object images}
In existing referring expression comprehension tasks, the target object always appears in the given image.
This is, however, unrealistic and uncommonly happens when we develop some glass-devices or autonomous robots that move around in scenes, capture images and search for the referred objects.
Therefore our annotation includes images where the referred object does not appear in them in the video video clips.
This imposes the referring expression comprehension models on practical and ambitious experimental settings: the models are required to discriminate images that include the referred objects from other images in the sequence of images of the video clip.
Our dataset includes 295,530 images in total and 226,319 images (76.6\% of the total images) contains the annotated bounding box of the referred target object in them. We also confirmed that at least four images in a single video clip have a target object annotation.
The lack of the target object from some frames is common in egocentric videos and becomes one of the greatest challenges for the conventional image-based REC models because such models are trained with images that surely include the target objects.

\noindent
\textbf{Multiple similar objects in scenes}
In previous referring expression comprehension datasets, it sometimes happens that some object classes are unique in one image and hence easy to localize the referred object just from the object class.
We consider this is due to the lack of the diverseness of the same-class objects in the same scene: if there is only one mug in some room, it is easy to select a mug from other objects just from an object class name.
However, it is much more rigorous to localize a single mug cup on a shelf full of mug cups from referring expressions.
We therefore choose the objects when there are multiple same-class of objects for them in the video clip for further annotation.
We select temporal scenes of video clips that include many objects for annotation.
The average number of detected objects in images is 22.1, and the average number of the same class objects present in the images with the target object is 3.5 for all splits~\footnote{We used Detic for this statistics.}.

\subsection{Dataset creation}
\noindent
\textbf{Video clip extraction from Ego4D}
We firstly extracted images from all of the original Ego4d videos and then applied the object detection model of Detic~\cite{detic} trained with the LVIS~\cite{lvis} dataset to automatically detect object bounding boxes for the extracted images.
Based on the object detection results, we sampled video clips that include many detected objects in frames. We did this because detecting and localizing objects from images with multiple objects can be challenging. We also choose video clips that have motions, avoiding video clips with less motion.
We also avoided stereo videos for annotations because they have special depth treatments and hence require additional annotation costs in later steps.
The frames of the sampled video clips consist of 10 to 40 images.
For preserving a wide variety of Ego4D dataset, we ensured the extracted video clips cover the most of activity themes annotated in Ego4D.
The detected bounding boxes are also used to provide bounding box candidates of possible target objects for later human annotation on Amazon Mechanical Turk (MTurk).
We chose 2 frames per second image extraction for the annotation, considering the annotation cost for longer video clips and the ability of tracking the same object in video clips of the existing object tracking datasets, such as TAO~\cite{tao2020} where frames are sampled in 1 frame per second.

\noindent
\textbf{Human annotation}
We used Amazon Mechanical Turk (MTurk) for collecting annotations to massive scale extracted video clips of 12,038.
We asked MTurk workers for choosing the same object in frames, editing the detected bounding box of the tracked object, and writing the referred expressions for the target object.
For this purpose, we developed an interactive visualization website that presents images that include candidates of bounding boxes from Detic drawn from the sampled video clips.
Workers are asked to select the same object in images and write down a referring expression to specify the object.
They are also asked to edit the bounding box of the target object by clicking the edges of them to fit the object.
Workers are asked to compose a referential expression that is enough detailed to localize the target object
by observing all frames in the video clips.
We further collected supplementary annotations of the referred object.
The further details of the video clip selection and annotation process are in S.M.

\begin{table}[t]
\begin{center}
	\small\begin{tabular}{lccc}
    \toprule
Split & \# Clips & \# Images & \# Images with BBox\\
\midrule
Train & 9,172 & 225,500 & 173,183 \\
Val.  & 1,549 & 38,470  & 29,322 \\
Test  & 1,317 & 31,560  & 23,814 \\
\bottomrule
	\end{tabular}
    \caption{
        Dataset statistics.
    }
    \label{table:dataset_stat}
\end{center}
\vspace{-1em}
\end{table}

\subsection{Dataset statistics}
We finally gathered annotations on 12,038 video clips. The total length of the video clip is 147,765 seconds and the averaged length is 12.3 seconds. Each clip has two different referring expression writings for one annotated object.
The Ego4D dataset has its own video clips for episodic memory, hands and objects and audio-visual diarization \& social tasks. We assumed our annotation serves supplementary roles for these existing tasks.
Therefore, for dataset splitting, we followed these existing splits as much as possible, namely, the Forecasting + Hands \& Objects (FHO) splitting. For some video clips without FHO, we follow Episodic Memory(EM) splitting. The remaining videos are for the training set.
Table~\ref{table:dataset_stat} presents the statistics for each split. We make sure that clips sampled from the same video are assigned to the same split.
The further detailed dataset statistics, including human accuracy of the REC task, and construction details of the annotations are in the S.M.

\section{Model}
\label{sec:baseline_model}

\subsection{Referring expression comprehension models}

We first apply the conventional image-based referring expression comprehension models of MDETR~\cite{mdetr} and OFA~\cite{ofa} for a sequence of images from video clips.

\noindent
\textbf{MDETR}
MDETR is an end-to-end text-modulated detector based on the DETR~\cite{detr}, state-of-the-art detection frameworks.
MDETR archived high performance on the REC benchmark, such as RefCOCO/+/g~\cite{refcoco, refcocog}.
It uses the soft token prediction to ground parts of textual expressions and detected regions in images through N learnable embeddings, called object queries, given to the MDETR decoder. Each bounding box prediction is also paired with a special token that represents that the bounding box is not grounded to the given textual phrase.
For prediction of each token $t^{*}_n$ of the referred expression with $L$ tokens, the MDETR decoder derives the probability $s^{i}_n$ that $i$-th token is paired to $n$-th object query as 
$s_n^i = \frac{\exp t_n^i}{\sum_{j=1}^{L+1} \exp{t_n^j}}$.
Here $i$ is a whole number and $1 \leq i \leq L+1$. $s_n^{L+1}$ is the special token prediction for ``no object'', of the $n$-th object query. We use $(1 - s_n^{L+1})$ as the confidence score for the bounding box prediction from the $n$-th object query.

\noindent
\textbf{MDETR with all images}
In contrast to existing REC datasets, some image frames in video clips don't include the referred objects in the RefEgo dataset.
When we train MDETR only with images that include the referred object, MDETR tends to predict bounding boxes with high confidence even if there are no referred objects in the images.
To assign low confidence scores for no referred object images, we trained a MDETR model with all images by treating all predicted bounding boxes as negative samples for images without referred objects during training.

\noindent
\textbf{MDETR+BH}
For better confidence scores to discriminate images without referred objects, we also expand the existing MDETR and add an additional object query to the existing prediction heads of MDETR. This additional object query is combined with a binary head (BH) to perform the binary classification trained with the binary cross-entropy loss by determining whether referred objects are in images or not.

\noindent
\textbf{OFA}
OFA~\cite{ofa} is the unifying architecture for various vision and language tasks, e.g., image captioning, visual question answering and visual grounding.
Unlike existing object detection models, OFA predicts a bounding box by directly determining the region position in $\braket{x1, y1, x2, y2}$-order with an autoregressive language model prediction. This nature, however, makes it difficult to obtain the ``confidence score'' for the predicted bounding box. We use the prediction probability of the sequence of $\braket{x1, y1, x2, y2}$ tokens as ``confidence score'' for the confidence of the prediction.
Similar to the original MDETR, we used extracted images that include the bounding box annotations for training.

\noindent
\textbf{No-referred-object images detection}
In RefEgo, models are required to discriminate images that don't contain the referred object.
For this purpose, we use the confidence scores of the predicted bounding boxes by simply assigning a threshold on it to determine whether images include the referred object or not.

\subsection{Object tracking}
\label{sec:object_tracking}

Unlike conventional referring expression comprehension on 2D images, RefEgo is a video-based REC dataset where an object is localized through the video frames.
Images in video frames often become worse because of motion blur and occlusion. Therefore it is difficult for REC models to consistently find the referred object in all image frames in sequence. In addition, because of the temporal movement of objects in videos, REC models often detect totally different objects in some frames, resulting in inconsistent object tracking in video frames.
To reduce inconsistent localization across video frames, We took the tracking-by-detection approach with attaching a tracking algorithm to the image-based REC models.
We applied ByteTrack~\cite{bytetrack} for the results from MDETR prediction.
ByteTrack is a state-of-the-art tracking algorithm that tracks objects based on the overlap between adjacent frames.
Because both the camera and objects can be in motion, ByteTrack calculates the overlap after predicting positions in the next frame with the Kalman filter.
We first extracted 30 frame per second images from the RefEgo video clips and obtained the MDETR predictions on them. We used ByteTrack for these MDETR prediction results and obtained the candidates of multiple tracked objects.
We then introduced a simple heuristics to score the sequences of bounding boxes of the tracked objects with the confidence score from MDETR, and update the predicted bounding boxes from MDETR if the confidence score of the tracked bounding boxes are higher than the original MDETR confidence score.
We consider this heuristics-based approach serves as the baseline of the object tracking over referring expression comprehension models.
We didn't apply object tracking for OFA results because OFA always predicts a single object in each frame. Thus there are no chances for object tracking to select better bounding boxes.
Please see S.M. for further details of the implementation and hyper parameters of the object tracking over referring expression comprehension models.

\begin{figure}[t]
    \centering
    \includegraphics[width=6cm]{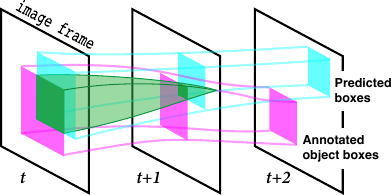}
    \caption{Image frames and STIoU overview. STIoU is defined with the intersection area (green) over the sum of the predicted (purple) and annotated (cyan) object boxes through time frames.}
    \label{fig:STIoU}
\end{figure}

\begin{table*}[t]
\begin{center}
    \footnotesize\begin{tabular}{lcccccccccc}
    \toprule
    & \multicolumn{5}{c}{RefEgo Val}  & \multicolumn{5}{c}{RefEgo Test} \\
    \cmidrule(lr){2-6}\cmidrule(lr){7-11}
    & \multicolumn{3}{c}{All images} & \multicolumn{2}{c}{Images w/ targets}  & \multicolumn{3}{c}{All images} & \multicolumn{2}{c}{Images w/ targets} \\
     \cmidrule(lr){2-4}\cmidrule(lr){5-6}\cmidrule(lr){7-9}\cmidrule(lr){10-11}
Model                & mSTIoU & mIoU+n & mAP@50+n & mIoU & mAP@50 & mSTIoU & mIoU+n & mAP@50+n & mIoU & mAP@50 \\
\midrule
\multicolumn{11}{l}{~~~~~\textit{From Pretrained}} \\
OFA           & 33.9 & 44.8 & 47.8 & \textbf{53.2} & \textbf{58.4} & 32.9 & 44.9 & 47.8 &  \textbf{51.9} & \textbf{56.8} \\
MDETR         & 36.2 & 42.1 & 47.9 & 46.0 & 53.3 & 35.4 & 42.4 & 48.0 & 45.0 & 52.3   \\
~~~~+Object tracking & 36.3 & 42.2 & 47.8 & 46.1 & 53.4 & 35.5 & 42.4 & 48.1 & 45.2 & 52.4  \\
MDETR (all)   & 37.2 & 45.6 & 51.2 & 45.0 & 52.3 & 36.1 & 45.0 & 50.5 & 44.0 & 51.1  \\
~~~~+Object tracking & 37.5 & 45.5 & 51.2 & 45.3 & 52.6 & 36.5 & 45.1 & 50.7 & 44.1 & 51.3   \\
MDETR+BH (all)& 37.5 & \textbf{46.3} & \textbf{52.0} & 45.2 & 52.6 & 36.5 & 45.6 & 51.0 & 45.4 & 52.7 \\
~~~~+Object tracking & \textbf{37.9} & 46.1 & 51.9 & 45.4 & 52.9 & \textbf{36.9} & \textbf{45.7} & \textbf{51.1} & 45.7 & 53.0 \\
\midrule
\multicolumn{11}{l}{~~~~~\textit{From RefCOCOg}} \\
OFA$^{\dagger}$      & 16.9 & 30.2 & 30.8 & 30.0 & 29.6   & 15.4 & 28.9 & 29.3 & 27.8 & 27.1 \\
OFA$^{\ddagger}$     & 32.7 & 44.5 & 47.7 & \textbf{52.3} & \textbf{57.7}   & 31.7 & 44.4 &  47.5 & \textbf{51.0} & \textbf{56.2} \\
MDETR$^{\dagger}$    & 17.4 & 27.4 & 28.3 & 25.1 & 25.2   & 15.4 & 25.6 & 26.4 & 22.9 & 22.8 \\
~~~~+Object tracking & 17.5 & 27.3 & 28.3 & 25.2 & 25.2   & 15.5 & 25.6 & 26.3 & 23.0 & 22.9 \\
MDETR$^{\ddagger}$   & 36.6 & 42.0 & 47.6 & 46.6 & 53.6   & 35.8 & 41.4 & 46.8 & 45.8 & 52.5 \\
~~~~+Object tracking & 36.7 & 41.9 & 47.5 & 46.7 & 53.7   & 35.9 & 41.3 & 46.7 & 45.9 & 52.7 \\
MDETR$^{\ddagger}$ (all)          & 37.9 & 45.3 & 50.9 & 46.4 & 53.5   & 37.2 & 45.0 & 50.4 & 45.7 & 52.6 \\
~~~~+Object tracking & 38.2 & 45.3 & 50.9 & 46.6 & 53.8   & 37.5 & 45.0 & 50.4 & 45.9 & 52.9 \\
MDETR+BH$^{\ddagger}$ (all)       & 37.5 & \textbf{46.1} & \textbf{51.6} & 46.4 & 53.6   & 36.9 & \textbf{45.7} & \textbf{51.1} & 45.7 & 53.0 \\
~~~~+Object tracking & \textbf{38.4} & 46.0 & \textbf{51.6} & 46.8 & 54.1   & \textbf{37.6} & 45.4 & 51.0 & 46.0 & 53.4 \\
\bottomrule
	\end{tabular}
    \caption{
        Experimental results on RefEgo validation and test sets.
        (${\dagger}$)~: the off-the-shelf RefCOCOg model performance.
        ($\ddagger$)~: models are trained with RefEgo from the off-the-shelf  RefCOCOg model. Other models are trained with RefEgo from the pretrained checkpoints.
    }
    \label{table:experiments_main_val}
\end{center}
\vspace{-1em}
\end{table*}

\section{Experiments}
\label{sec:experiments}

\subsection{Evaluation metric}

In 2D referring expression comprehension, the mean intersection-over-union overlap (\textbf{mIoU}) is commonly used for the quality of the target object selection.
Following the widely used metric for referring expression comprehension~\cite{refcocog,refcoco}, we count $\mathrm{IoU}>0.5$ cases for positive cases and otherwise negative cases for \textbf{AP@50}.
\textbf{AP@50} is not sensitive to the details of the shape of the predicted bounding boxes but  the selection of the objects from other similar objects.
These metrics, however, are applicable only when the target object is presented in the images.
We apply the traditional \textbf{mIoU} and \textbf{AP@50} for images that include the annotated target object bounding box, ignoring image frames that don't contain the referred objects.

As the video-based object tracking task, the target object can be invisible or out-of-frame in the sequence of image frames.
This is a major difference of this task compared with the existing object tracking datasets.
We therefore expand the existing mean IoU metric for the video-based evaluation including frames that don't contain the target object: mean Spatio-Temporal IoU (mSTIoU) and mean IoU with negative prediction (mIoU+n).

Suppose we have a single video clip that consists of $\mathcal{N}$ frames in the frame-per-second evaluated.
For $i$-th image frame in $\mathcal{N}$ frames include an annotated bounding box $t_i$ where its area size is given by $|t_i|$.
Among $\mathcal{N}$ frames, $\mathcal{M}$ frames include the target object and hence $|t_i|>0$ while $|t_i|=0$ for the remaining $\overline{\mathcal{M}} \cap \mathcal{N}$ frames (here $0 \leq |\mathcal{M}| \leq |\mathcal{N}|$).
For each image frame, models predict a bounding box $p_i$ that may become size-0 ($|p_i| \geq 0$), suggesting that the image frame doesn't contain the referred object in it for $|p_i| = 0$ case.
The conventional mIoU only for images that include the target object is:
\begin{equation}
    \mathrm{mIoU} = \frac{1}{|\mathcal{M}|}\sum_{\mathcal{M}}\frac{|p_i \cap t_i|}{|p_i \cup t_i|}.
\end{equation}
\label{eq:iou}
This is an image-wise metric, ignoring images that don't contain the target object ($\overline{\mathcal{M}} \cap \mathcal{N}$).

We introduce the Spatio-Temporal IoU (STIoU) as the multi-frame summation of the intersection and union of IoUs over a single  video clip of $\mathcal{N}$ frames:
\begin{equation}
    \mathrm{STIoU} = \frac{\sum_{\mathcal{N}} |p_i \cap t_i|}{\sum_{\mathcal{N}} |p_i \cup t_i|}.
\end{equation}
STIoU satisfies $0 \leq \mathrm{STIoU} \leq 1$ where $\mathrm{STIoU} = 1$ for the exact math in all frames while  $\mathrm{STIoU} = 0$ for complete mismatch of all annotated and predicted bounding boxes. When $|t_i|=0$, STIoU is not penalized if $|p_i| = 0$ while it decreases due to the larger denominator of $|p_i \cup t_i|=|p_i|$ if $|p_i| > 0$.
We use mean STIoU (mSTIoU) where the STIoU is calculated in each video clip and then we take the mean of STIoU inside the validation and test splits.

For Eq.~\ref{eq:iou}, we cannot replace $\mathcal{M}$ with $\mathcal{N}$ without any assumptions because $|p_i \cup t_i|$ can become 0 for $|t_i|=0$ case.
However, for $|t_i|=0$ case, $|p_i \cap t_i|$ is always 0 for any $|p_i|$.
Therefore, we extend the conventional IoU metric for $\mathcal{N}$ under an assumption that for $|t_i| \to 0$ and $|p_i| \to 0$ case, $\frac{|p_i \cap t_i|}{|p_i \cup t_i|} \to 1$.
We call this \textbf{IoU+n} as IoU with the negative prediciton.
\textbf{IoU+n} can be expressed in the following simple form:
\begin{equation}
\begin{split}
    &\mathrm{IoU+n} = 
    \begin{cases}
    \frac{|p_i \cap t_i|}{|p_i \cup t_i|} & (|p_i|>0~~\mathrm{or}~~|t_i|>0)\\
    1 & (|p_i|=0~~\mathrm{and}~~|t_i|=0)
    \end{cases}
\end{split}
\end{equation}
Unlike simple IoU, IoU+n includes the detection of images without the target object.
Similar to IoU, we take mean of IoU+n for images and video-clips (mIoU+n), and
similar to AP@50, we also introduce \textbf{AP@50+n} where we count the $\mathrm{IoU+n}>0.5$ cases.
Fig.~\ref{fig:STIoU} visualize the STIoU metric as an IoU extension for time sequences.

\begin{figure}[t]
    \centering
    \includegraphics[width=8.5cm]{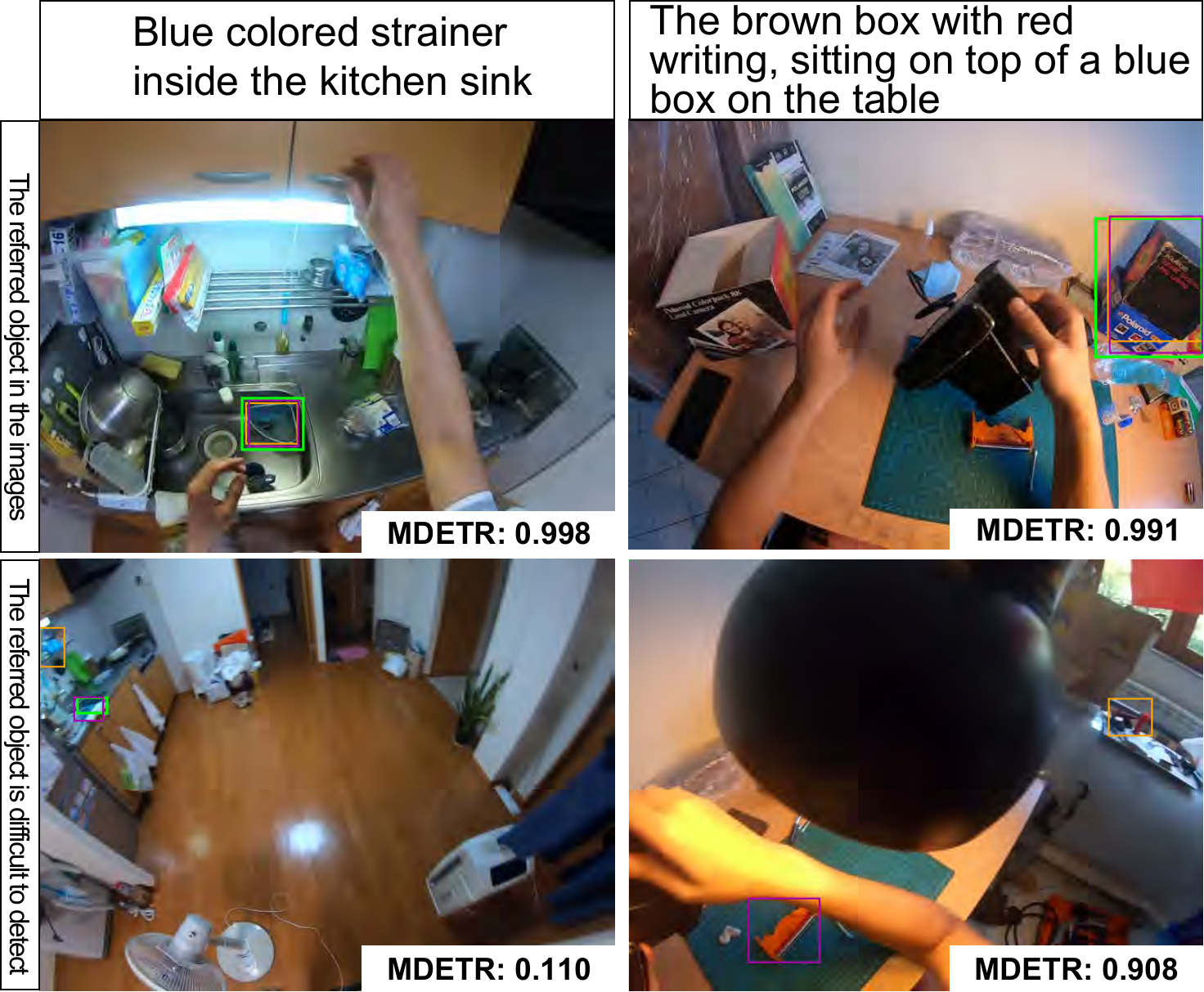}
    \caption{Example bounding box annotated in \textcolor{green}{green}, predicted by OFA$^{\ddagger}$ in \textcolor{orange}{orange} and MDETR$^{\ddagger}$ (all) in \textcolor{purple}{purple} in two images in two columns from the same video clips.
    When the referred object doesn't exist in the images (below), both OFA$^{\ddagger}$ and MDETR$^{\ddagger}$ models detect bounding boxes of other objects.}
    \label{fig:no_annnotaed_boxes}
    \vspace{-0.5em}
\end{figure}

\begin{figure}[t]
\begin{tabular}{cccc}
\begin{minipage}{.3\textwidth}
    \includegraphics[width=4cm]{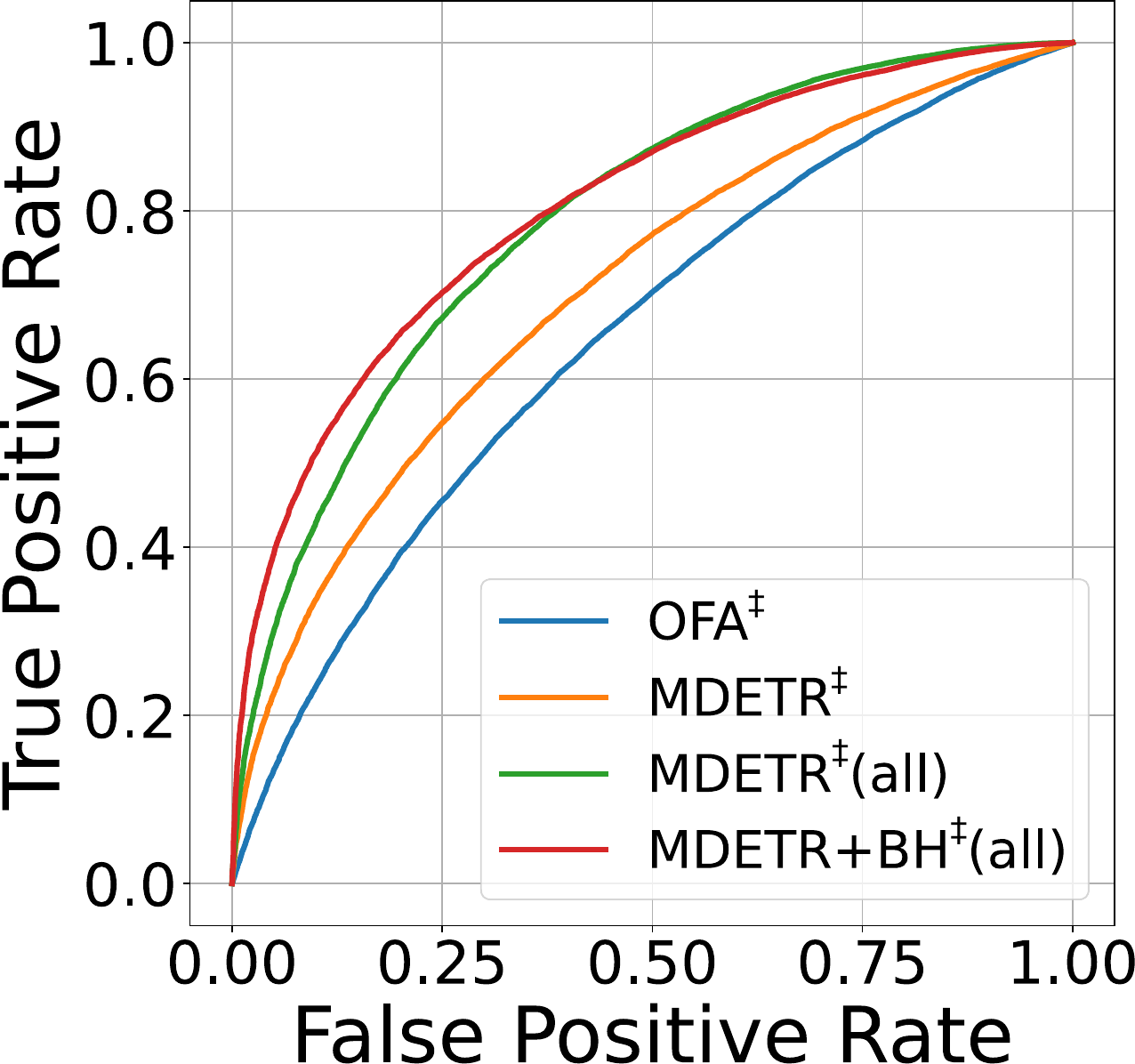}
    \vspace{0.5em}
\end{minipage}
\begin{minipage}{.2\textwidth}
\begin{center}
\vspace{-1em}
\hspace{-1.5cm}
	\scriptsize\begin{tabular}{lc}
    \toprule
Model                & AUC \\
\midrule
(Chance rate)          & 50.0 \\
\midrule
OFA$^{\ddagger}$  & 64.9 \\
MDETR$^{\ddagger}$ & 70.7 \\
MDETR$^{\ddagger}$ (all)   & 78.8 \\
MDETR+BH$^{\ddagger}$ (all)   & 80.5 \\

\bottomrule
	\end{tabular}
    \label{table:experiments_auc}
\end{center}
\end{minipage}
\end{tabular}
\caption{
\textbf{Left}: ROC curves for detection of the no-referred-object images.
\textbf{Right}: AUC for the prediction whether the referred objects are in images or not.
}
\label{fig:roc curve}
\end{figure}

\begin{figure*}[t]
    \centering
    \includegraphics[width=17cm]{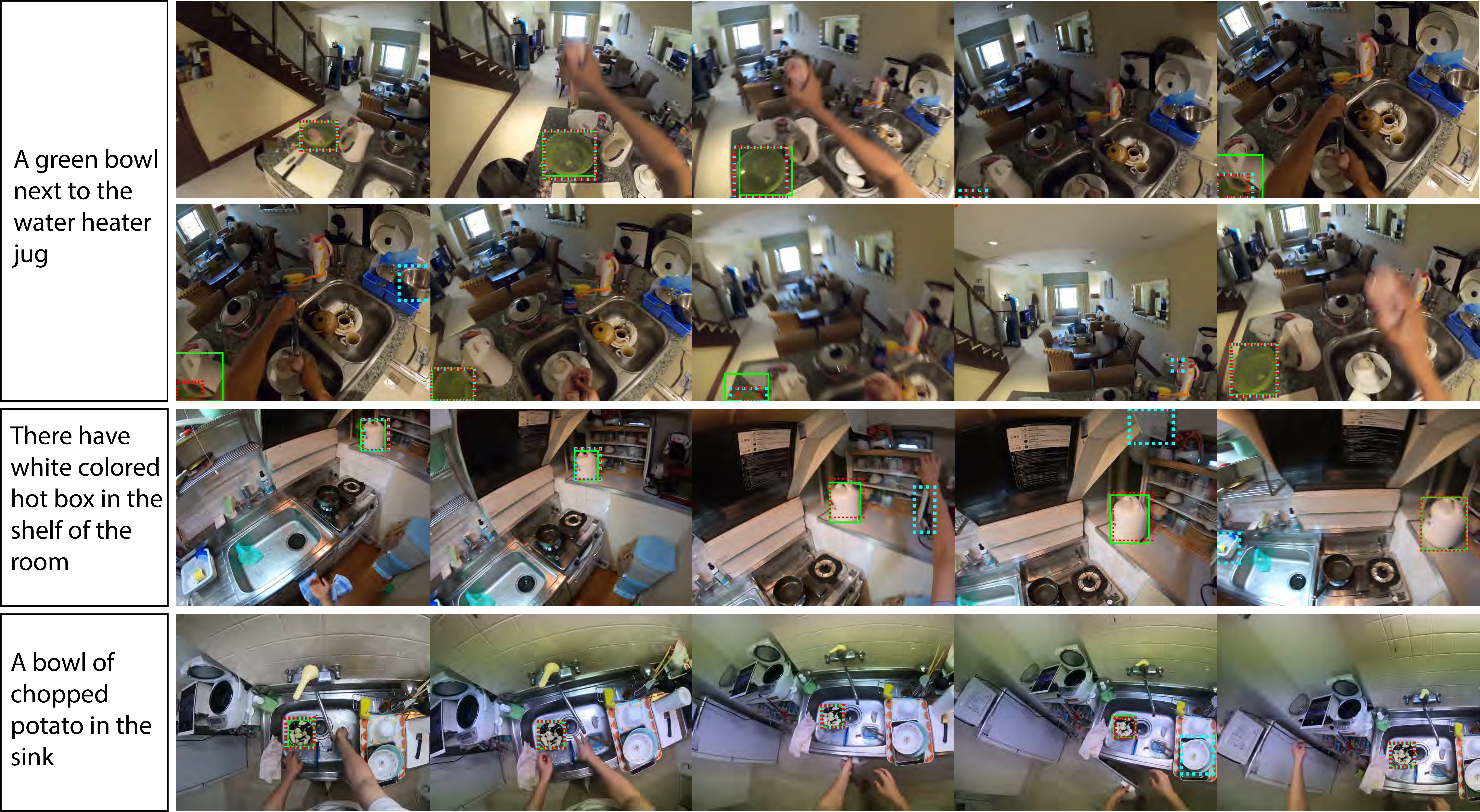}
    \caption{
    Qualitative analyses for three video clips. Selective images from left-top (past) to right bottom (future). The green, dotted red and dotted cyan bounding boxes are \textcolor{green}{annotated}, \textcolor{red}{predicted by ByteTrack} and \textcolor{cyan}{predicted by REC} with the top-1 confidence score, respectively. The texts at the left of images are the referring expressions.
    }
    \label{fig:bytetrack}

\end{figure*}

Here, the introduced STIoU and IoU+n are video-wised evaluation metrics. STIoU is penalized for predicting finite-sized bounding boxes for images without target objects while IoU+n is rewarded for predicting images without the targets. Among these, \textbf{STIoU is the prime video-wised metric of the video-based referring expression comprehension} because it is based on the entire video-clip coverage of the referred object. It is notable that IoU+n is too sensitive to the corner cases where the small fragment of the referred object is presented at the edge of the image frame and detecting such fragments becomes a subtle but critical problem for IoU+n. STIoU is robust to such corner cases because STIoU isn't affected too much by small bounding boxes of the predictions and annotations.


\subsection{Video-based REC in RefEgo}

We used the state-of-the-art pretrained REC models of OFA-Large~\cite{ofa} and MDETR~\cite{mdetr} for our experiments. For MDETR, EfficientNet-B3~\cite{efficientnet} is used the visual backbone network in experiments.
We prepared OFA, MDETR and MDETR (all) models.
OFA and MDETR models use extracted image frames that include the annotated bounding boxes
while MDETR (all) uses all image frames including images that do not contain the referred object bounding box as described in Sec.~\ref{sec:baseline_model}.
We trained OFA and MDETR models with RefEgo from the pretrained checkpoints of these models.
We also prepare OFA$^{\ddagger}$ and MDETR$^{\ddagger}$ models from the off-the-shelf models of OFA$^{\dagger}$and MDETR$^{\dagger}$ trained with RefCOCOg~\cite{refcocog}.

Table~\ref{table:experiments_main_val} presents the performance of the REC models in RefEgo validation and test sets.
The current state-of-the-art referring expression comprehension models somehow successfully localize objects in images when the target object is surely in the given image as the OFA models exhibit strong performances in AP@50 of the widely used REC metrics in the validation set.
The MDETR$^{\ddagger}$ (all) and MDETR+BH$^{\ddagger}$ (all) models, however, achieve better performance in the all images metrics of mSTIoU, mIoU and AP@50+n. This suggests that the confidence scores of these models are more effective in determining the presence or absence of the referred object in the image compared to other models, assuming that this is because the MDETR$^{\ddagger}$ (all), trained with all images, learns the images that do not include the referred object through contrastive learning. Overall, OFA models are good at predicting accurate bounding boxes when the referred objects are in the images while MDETR models are good at predicting both the bounding boxes and discriminating images without the referred objects.

We also applied ByteTrack-based object tracking for MDETR models as described in Sec.~\ref{sec:object_tracking}. In Table~\ref{table:experiments_main_val}, we observed the object tracking slightly improves performance especially in mSTIoU and mIoU while it is less effective for mIoU+n.
We noticed that the performance with mIoU+n sometimes slightly degrades with object tracking because of the lower discrimination of images without the referred objects.
We will take a close look in the quantitative analyses on the comparison between REC and object tracking results.

\subsection{Discriminating images without referred objects}

It is a difficult task to discriminate images without the referred objects from image frames of the video clips.
When the referred objects are out-of-images or not visible, the models have a tendency to predict bounding boxes on objects that appear similar to the referenced expression but are not the correct ones.
Fig.~\ref{fig:no_annnotaed_boxes} presents the predictions of bounding boxes for both models for cases where the target object is in the image (top) and the referred object is difficult to detect or isn't visible in the images due to occlusion caused by other objects.(bottom).
We further investigate how accurately REC models discriminate images without referred objects by metrics used for binary classification evaluations under various thresholds on the top-1 bounding box confidence scores.
Fig.~\ref{fig:roc curve} presents AUC (Area Under the ROC Curve) for ROC.
MDETR$^{\ddagger}$ (all) outperforms OFA$^{\ddagger}$ and MDETR$^{\ddagger}$ models in AUC, suggesting better performance in determining images that include the referred object. The binary head of MDETR+BH$^{\ddagger}$ (all) further contributes discriminating images without referred objects.

\begin{table*}[t]
\begin{center}
    \footnotesize\begin{tabular}{lcccccccccc}
    \toprule
    & \multicolumn{5}{c}{RefEgo Val}  & \multicolumn{5}{c}{RefEgo Test} \\
    \cmidrule(lr){2-6}\cmidrule(lr){7-11}
    & \multicolumn{3}{c}{All images} & \multicolumn{2}{c}{Images w/ targets}  & \multicolumn{3}{c}{All images} & \multicolumn{2}{c}{Images w/ targets} \\
     \cmidrule(lr){2-4}\cmidrule(lr){5-6}\cmidrule(lr){7-9}\cmidrule(lr){10-11}
Model                & mSTIoU & mIoU+n & mAP@50+n & mIoU & mAP@50 & mSTIoU & mIoU+n & mAP@50+n & mIoU & mAP@50 \\
\midrule
\multicolumn{5}{l}{\textit{Single object of same-class (easy)}} \\
OFA$^{\ddagger}$           & 36.5 & 48.0 & 51.4 & 58.2 & 63.9 & 35.6 & 48.2 & 51.6 & 56.6 & 62.4 \\
MDETR+BH$^{\ddagger}$ (all)& 43.9 & 51.9 & 58.1 & 52.3 & 60.3 & 42.4 & 51.3 & 57.2 & 51.3 & 59.2 \\
~~~~+object tracking       & 44.9 & 51.9 & 58.2 & 52.8 & 60.9 & 43.1 & 51.1 & 57.2 & 51.5 & 59.4 \\
\midrule
\multicolumn{5}{l}{\textit{Multiple objects of same-class (hard)}} \\
OFA$^{\ddagger}$           & 29.3 & 41.4 & 44.5 & 47.2 & 52.3 & 28.8 & 41.5 & 44.5 & 46.7 & 51.6 \\
MDETR+BH$^{\ddagger}$ (all)& 32.0 & 41.2 & 46.0 & 41.3 & 47.8 & 32.8 & 40.9 & 45.8 & 41.7 & 48.3 \\
~~~~+object tracking       & 32.8 & 41.5 & 46.6 & 41.6 & 48.4 & 33.6 & 41.2 & 46.5 & 42.0 & 49.0 \\
\bottomrule
\multicolumn{5}{l}{\textit{Static object (easy)}} \\
OFA$^{\ddagger}$            & 32.9 & 44.7 & 48.0 & 52.6 & 58.2 & 32.9 & 45.5 & 48.8 & 51.9 & 57.4 \\
MDETR+BH$^{\ddagger}$ (all) & 37.8 & 46.6 & 52.2 & 46.9 & 54.3 & 37.8 & 46.6 & 52.2 & 46.8 & 54.2 \\
~~~~+object tracking        & 38.6 & 46.4 & 52.2 & 47.3 & 54.9 & 38.5 & 46.4 & 52.1 & 47.0 & 54.6 \\
\midrule
\multicolumn{5}{l}{\textit{Moving object (hard)}} \\
OFA$^{\ddagger}$            & 31.3 & 43.1 & 46.0 & 50.4 & 54.9 & 25.9 & 38.8 & 41.3 & 45.9 & 50.2 \\
MDETR+BH$^{\ddagger}$ (all) & 35.9 & 43.5 & 48.0 & 43.8 & 49.6 & 32.4 & 40.7 & 45.6 & 40.6 & 46.7 \\
~~~~+object tracking        & 36.9 & 43.6 & 48.2 & 44.1 & 50.0 & 33.1 & 40.6 & 45.7 & 41.1 & 47.4 \\
\bottomrule
	\end{tabular}
    \caption{
        The performance difference due to the  the \textit{object-class uniqueness} and
        \textit{referred object movement} during prediction.
        \textbf{Top}: single (easy) and multiple (hard) objects of the same-class in image frames.
        \textbf{Bottom}: static (easy) and moving (hard) referred object.
    }
    \label{table:experiments_detailed_static}
\end{center}
\vspace{-1em}
\end{table*}

\subsection{Qualitative analysis}

Figure~\ref{fig:bytetrack} shows the results of MDETR+BH$^{\ddagger}$ (all) and its object tracking counterparts.
In the first two rows of the images, the referred object of the green bowl often goes at the edge of the frames or even out-of-frames in the video clip, making it a challenging scenario for conventional object tracking methods.
However, we found that the ByteTrack-based object tracking mode, with the assistance of the REC results, was able to successfully combine the tracked object of the green bowl in many frames.
In the images in the third and fourth rows, the REC model makes incorrect bounding box predictions in the middle of these two video clips. However, the object tracking method successfully continues to track the same object.
These video clips include multiple objects of the same class, which may cause confusion for the REC models.

\subsection{Detailed performance analyses}
\label{app:Detailed model performance}

We present detailed analyses on the \textit{object-class uniqueness} and \textit{referred object movement}.
If there are multiple objects of the same class in one video clip, it becomes more difficult to track the identical object. Similarly, if the referred object is moved to another place during the video clip, it is challenging to precisely localize the referred object in images frames.
Based on the OFA$^{\ddagger}$ and MDETR+BH$^{\ddagger}$ (all) models, we derived the detailed scores for the following four cases in the RefEgo validation and test sets in Table~\ref{table:experiments_detailed_static}, confirming that the model performance degrades when the objects are moved or multiple similar objects exist in the scenes.

\section{Conclusion}
\label{sec:conclusion}

Based on the wide-variety world-wide first-person perception dataset of Ego4D,
we constructed the RefEgo dataset for the real-world and egocentric video-based referring expression comprehension.
This dataset is not only larger than existing REC datasets in terms of images with annotated bounding boxes, but it is also grounded on the real-world egocentric videos, making it a valuable and challenging task for precisely grounding natural languages in real-world contexts.
In experiments, we combined the REC and object tracking approaches to spatio-temporally localize referred objects even in challenging conditions, such as when the referred object goes out-of-frames in the middle of the clips or the REC model makes several incorrect predictions. This approach provides a precious baseline for first-person video-based REC datasets.

\paragraph{Limitations} We have annotated our dataset using images from the Ego4D first-person video dataset. As a result, our usage terminology and limitations for videos and images align with those of the original Ego4D dataset. Our dataset encompasses diverse video domains, including indoor and outdoor scenes. However, it is important to note that we haven't included videos from domains rarely found in the Ego4D dataset.

\paragraph{Acknowledgments} This work was supported by JSPS KAKENHI Grant Number 22K17983 and 22KK0184, and by JST PRESTO Grant Number JPMJPR20C2.

{\small
\bibliographystyle{ieee_fullname}
\bibliography{egbib}
}

\clearpage
\appendix
\section*{Appendix}

\section{Dataset Details}
We constructed the RefEgo video-based referring expression comprehension dataset based on the world-wide various-domains first-person videos of Ego4D.
Our dataset covers 5,012 videos of all Ego4D~\cite{Ego4D2022CVPR} videos.
The total length of video clips are more than 41 hours.
We summarize the statistics of our dataset in Table~\ref{table:detailed_stat}.
Each video clip has two referring expression annotations.
In the validation and test set, we also collect additional classification labels for the referred object in each video clip: the \textit{object-class uniqueness} and \textit{referred object movement}.
\textit{Object-class uniqueness} is the label whether there are any other objects of the same class with the referred object or not.
\textit{Referred object movement} is the label whether the referred object is moving/moved in the video clip or not.
These additional labels are used for detailed analyses in Sec.~\ref{app:Detailed model performance}.

\subsection{Video clips in RefEgo}

\begin{figure}[t]
\begin{tabular}{cccc}
\begin{minipage}{.3\textwidth}
    
    \centering
    \hspace{-1.65cm}
    \includegraphics[width=4cm]{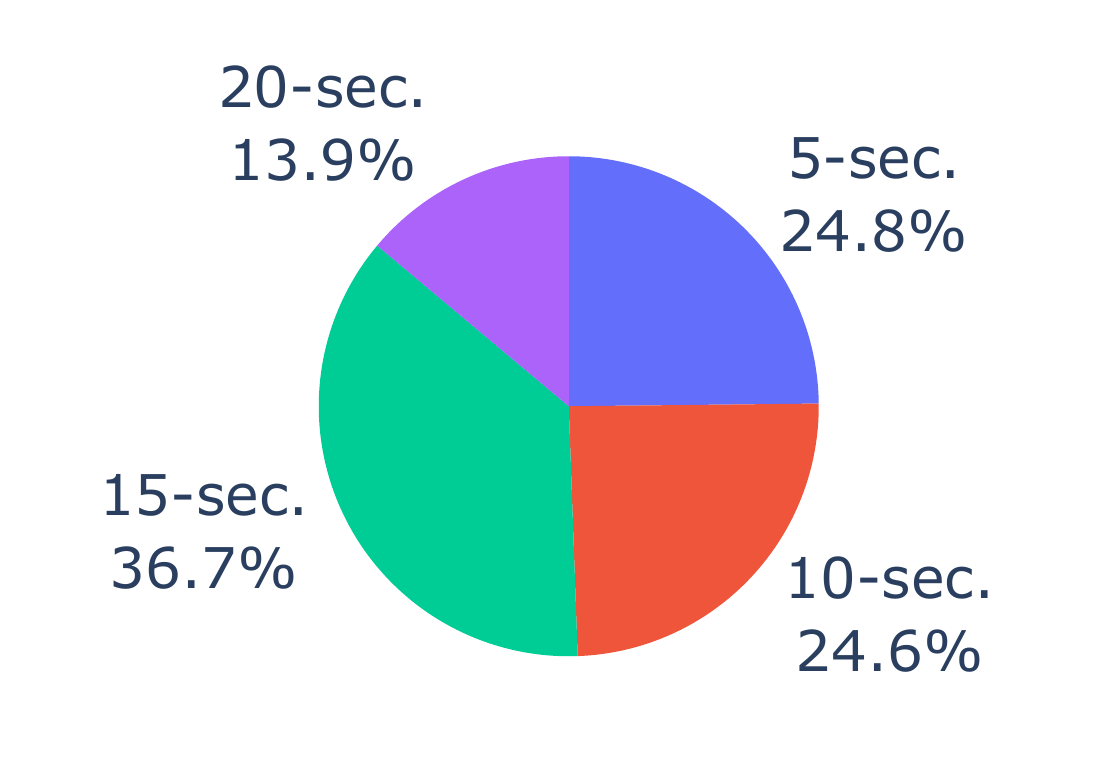}
\end{minipage}
\begin{minipage}{.2\textwidth}
    \centering
    \hspace{-2cm}
    \includegraphics[width=4cm]{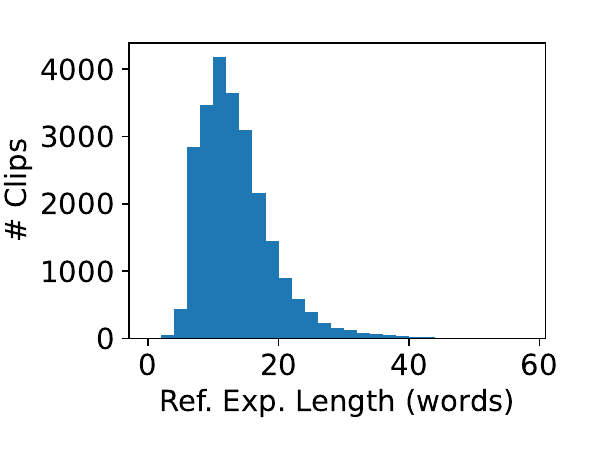}
\end{minipage}
\end{tabular}
\caption{
\textbf{Left}:  Ratio of video clip length.
\textbf{Right}: Length of referring expression.
}
\label{fig:clip_length_dist}
\end{figure}

\begin{figure}[t]
\begin{tabular}{cccc}
\begin{minipage}{.3\textwidth}
    
    \centering
    \hspace{-1.65cm}
    \includegraphics[width=4cm]{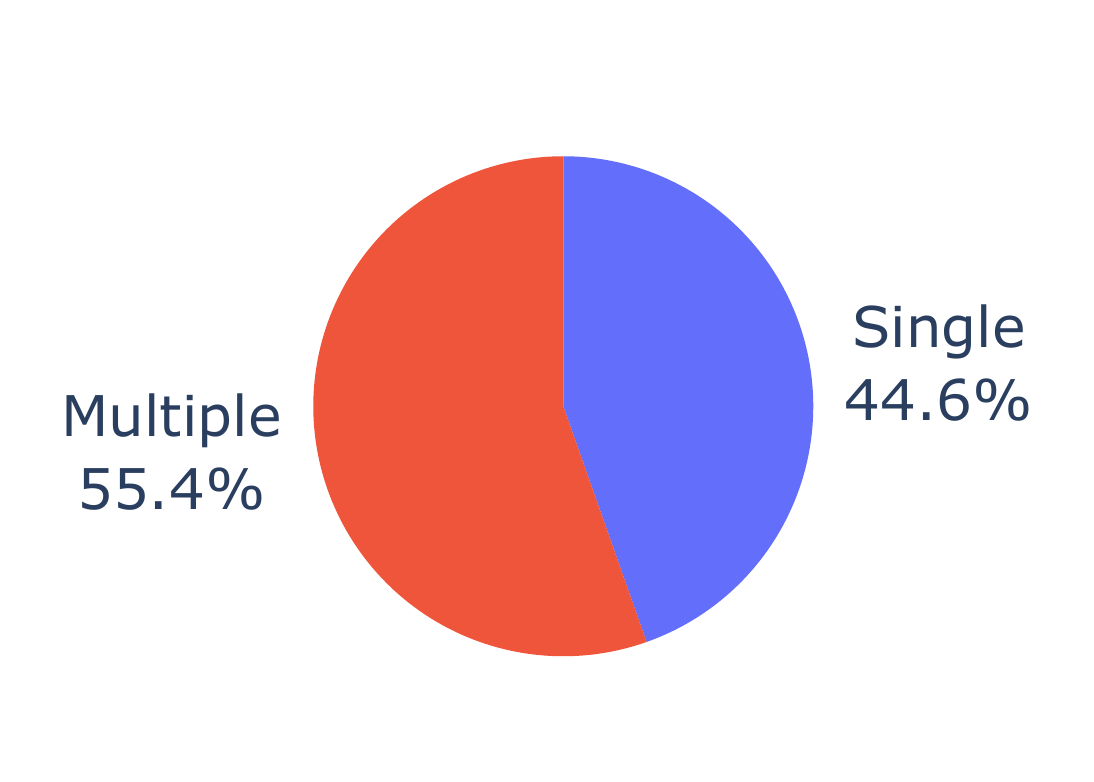}
\end{minipage}
\begin{minipage}{.2\textwidth}
    \centering
    \hspace{-2cm}
    \includegraphics[width=4cm]{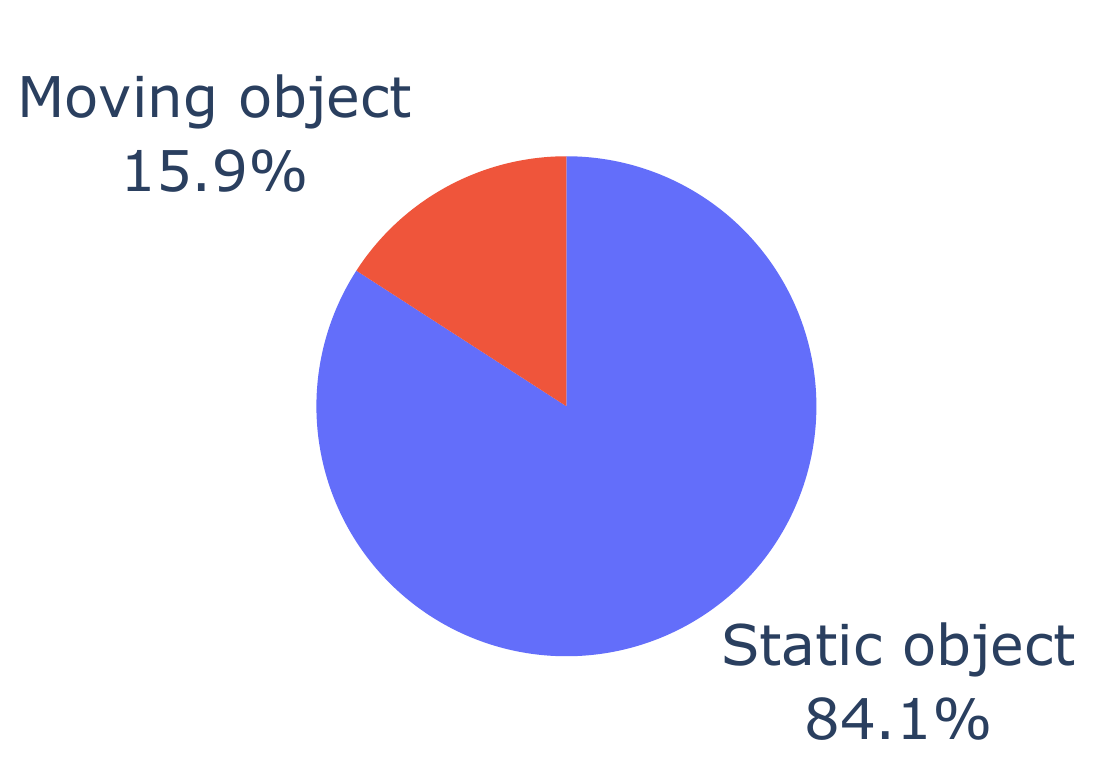}
\end{minipage}
\end{tabular}
\caption{
\textbf{Left}:  Ratio of the single object of the same-class and the multiple object of the same-class in image frames.
\textbf{Right}: Ratio of the static referred object and the moving referred object.
}
\label{fig:clip_detailed_object_type}
\end{figure}

\begin{figure}[t]
    \centering
    \includegraphics[width=6cm]{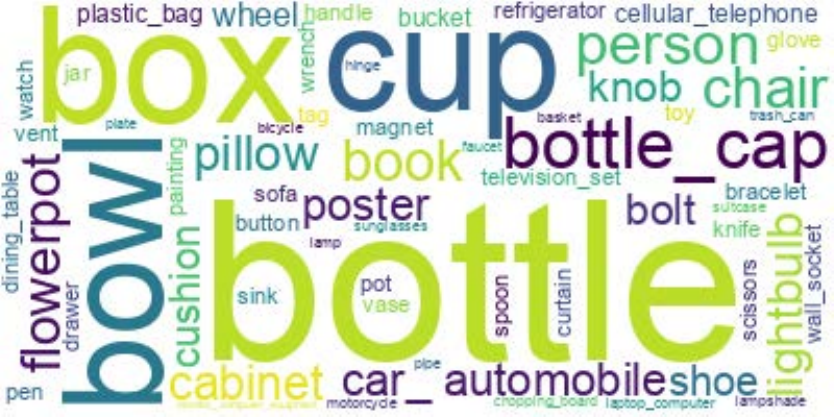}
    \caption{Frequently seen object categories in RefEgo by Detic (LVIS).}
    \label{fig:detic labels wordcloud}
\end{figure}

\begin{figure}[t]
    \centering
    \includegraphics[width=6cm]{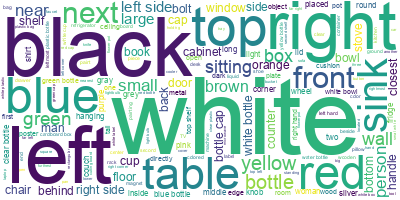}
    \caption{Frequently seen words in referential expression of RefEgo.}
    \label{fig:ref_wordcloud}
\end{figure}

In the RefEgo dataset, we firstly tried to cover as much as videos in Ego4D in order to follow a variety of topics.
However, we soon notice that some videos are not suitable for the referring expression comprehension task because they have a considerably small amount of detected objects in image frames. The variety of object classes is also limited.
Therefore we chose video clips in terms of the variety of objects.
Firstly we extracted two images per second from all images and applied the object detector Detic.
Then we chose the video clips where many classes of objects are in them according to the Detic results.
We also manually removed several videos that do not include many objects and hence are not suitable for the referring expression comprehension task (e.g., farm fieldwork or boarding on a leisure boat).
For the remaining videos, we sampled video clips that include as many objects as possible and they are in motion.
We notice there are some videos with very few motion of the view points (e.g., watching a movie).
We therefore sampled video clips in motion using both absolute difference of images and difference of the detected objects.
The length of the sampled video clips are chosen in $\{5,10,15,20\}$ seconds.
We present the ratio of each clip length and the distribution of the referring expression length in Fig.~\ref{fig:clip_length_dist}.
The video clip ratio of \textit{object-class uniqueness} and \textit{referred object movement} in the RefEgo validation and test set is presented in Fig.~\ref{fig:clip_detailed_object_type}.
Frequently seen object categories inferred by Detic are presented in Fig.~\ref{fig:detic labels wordcloud} and
frequently seen words in captions in Fig.~\ref{fig:ref_wordcloud}.

\begin{table}[t]
\begin{center}
	\small\begin{tabular}{lc}
    \toprule
\# Annotated video clips & 12,038 \\
\# Sampled Ego4D videos & 5,012 \\
Annotated FPS          & 2 \\
Total video length (sec) & 147,765 \\
Ave. video length (sec) & 12.3 \\
Ave. ref. exp. length (words) & 13.4 \\
\bottomrule
	\end{tabular}
    \caption{
        Detailed statistics for RefEgo dataset.
    }
    \label{table:detailed_stat}
\end{center}
\end{table}

\subsection{Dataset Annotation}
The annotation process via Amazon Mechanical Turk (MTurk) is performed by following two steps:
(i) object tracking and first referring expression attachment,
(ii) tracked object check and second referring expression attachment.
In the first step, we collected the tracking data for a single object in the video clip.
In the second step, we ask workers to check whether the same object is tracked for the results of the step-(i).
We also manually check the failed video clips and remove some of those clips if they are not suitable for our task.
Such filtered video clips are re-annotated through step-(i) again.

\paragraph{(i) object tracking and first referring expression attachment}
In object tracking annotation, our approach is the selection and correction from the boundary box candidates.
As written in the previous section, we automatically extracted object boundary boxes in each image with Detic.
We then present MTurk workers all images in the sampled video clips with the extracted boundary boxes for each image.
The all extracted boundary boxes serve as candidates for the single tracked object.
We recommend that workers view all images in the sampled clip first and then select a single boundary box for each image.
Workers can reshape bounding boxes to fit the target object if no boundary boxes are presented on the target object in some images.

For the target object class, we sampled one from the frequent object classes in the video clip depending on the LVIS class of Detic.
We present workers this target object class and ask them to find one of the objects and track it.
Note that we allow workers to track objects that are not in the target object class if they are unavailable or difficult to track in images.
We infer the tracked object labels based on the most frequent object labels after the entire annotation.

\paragraph{(ii) tracked object check and second referring expression attachment}

We present workers images where a single referred object with a bounding box is attached for each image and ask them to check whether the same object is tracked through the images.
\textbf{Here, to make sure the same objects are tracked in the dataset, object tracking annotations that do not pass the same object tracking test in step-(ii) are removed from the dataset or re-annotated in (i) step.} $5.8$\% annotations are marked as they do not track the same object in the video clips and hence removed or re-annotated. We re-annotate or remove these video clips. Note that this figure doesn't include the results of ``null workers'' in MTurk because results of ``null workers'' are removed and reassigned to other workers immediately upon they are found.
We also ask workers to compose an additional referring expression in addition to the auxiliary information labels of the object stationary and uniqueness in all instances in the validation and test sets.
Therefore, each video clip has two referred expression annotations.

\begin{figure*}[t]
    \centering
    \includegraphics[scale=0.38]{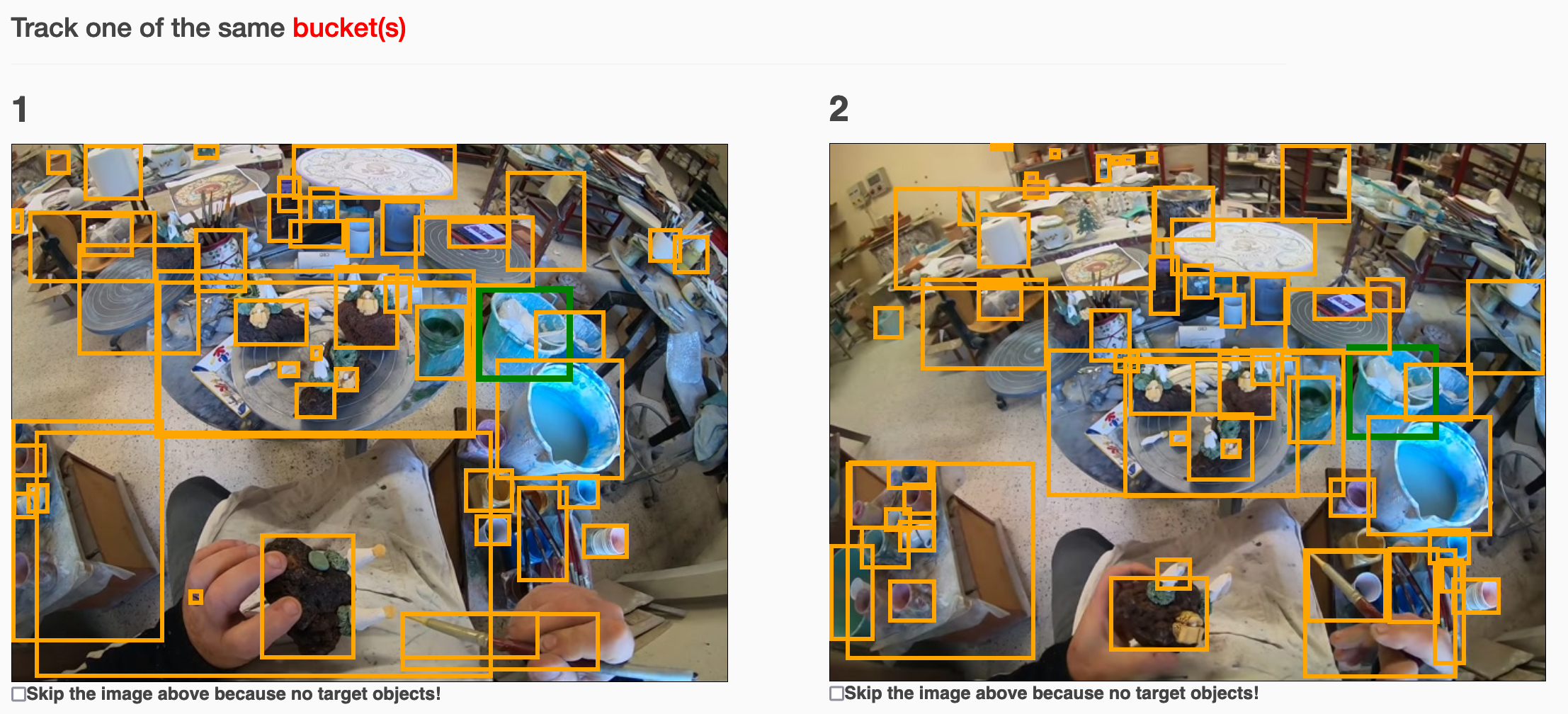}
    \caption{MTurk annotation website.}
    \label{fig:mturk_website}
\end{figure*}

After the step (ii), we quickly reviewed overall annotations and manually edited them if necessary.
We confirmed that we gathered two different referring expressions for each tracked object.
We present a sample annotation website on MTurk to adjust and select the bounding boxes in Fig.~\ref{fig:mturk_website}.

\begin{table}[t]
\begin{center}
    \footnotesize\begin{tabular}{lcccc}
    \toprule
    & \multicolumn{2}{c}{Val.} & \multicolumn{2}{c}{Test}  \\
Model                & mIoU & mAP@50 & mIoU & mAP@50 \\
\midrule
CG-SL-Att~\cite{cogrounding_2021_CVPR}  & 33.2 & \textbf{38.0} & 32.9 & \textbf{38.0} \\
DCNet~\cite{cao2022correspondence}      & \textbf{34.2} & 37.0 & \textbf{33.2} & 36.5 \\
\bottomrule
	\end{tabular}
    \caption{
        RefEgo val. and test sets (Images w/ targets).
    }
    \label{table:experiments_dcnet}
\end{center}
\vspace{-2em}
\end{table}

\section{Video-based REC baselines}
\label{app:Detailed model performance}

We also derive the scores with existing video-based REC models of Co-grounding network~\cite{cogrounding_2021_CVPR} and DCNet~\cite{cao2022correspondence}.
It is considerable that these models do not concentrate on the discrimination of images without target objects and therefore we train these models with images w/ targets.
Results are evaluated in image-based metrics of mIoU and mAP@50 in Table~\ref{table:experiments_dcnet}.
These models use alignments between frames that are not suitable when we apply these models for extracted frames that include the target objects.
It is notable that OFA and MDETR have a strong object detection ability from pretraining and hence achieve a good performance even in the detection setting.

\section{Object tracking implementation details}

For ByteTrack~\cite{bytetrack}, we used GIoU~\cite{Rezatofighi_2018_CVPR} for this similarity criteria to enable robust matching in motion videos. We set thresholds of the high and low scores for the detection boxes to 0.1 and -0.5, respectively. For object matching between adjacent image frames, the matching candidate object bounding boxes are allowed only when the REC confidence score (or objectness scores of MDETR) are greater than 0.9 and GIoU is greater than 0.9.
In addition, we apply NMS to predictions in each frame to reduce overlapped bounding boxes.
We summarize the hyper-parameters of ByteTrack in Table~\ref{table:object_track}.

We notice that the confidence scores from REC models are sufficiently reliable even when objects that are not well-tracked.
Therefore, we use a simple heuristic to merge the object tracking results with the original REC results.
We first assign the REC confidence score for the new confidence score of the candidate object bounding boxes.
Then, following the object tracking results, we obtained the averaged REC confidence score for the time-sequence of the bounding boxes that are sequenced by ByteTrack.
If an averaged confidence score is higher than the original score for some bounding boxes, we updated the confidence score with the averaged one.
By doing so, we can maintain the REC confidence score when it is sufficiently high while we can update the low REC confidence score for the bounding boxes when the bounding boxes for the same object in adjacent frames are sufficiently high.

\begin{table}[t]
\begin{center}
	\small\begin{tabular}{lc}
    \toprule
Track high threshold & 0.1 \\
Track low threshold  & -0.5 \\
Confidence Score threshold & 0.9 \\
GIoU Match threshold & 0.9 \\
\bottomrule
	\end{tabular}
    \caption{
        ByteTrack parameters.
    }
    \label{table:object_track}
\end{center}
\end{table}

\section{Human performance on RefEgo}
To compare the current model performance with those by the human experts, we provide the human performance for the test set of the RefEgo.
For this human expert test, we first sampled video clips in the test set and presented them in the same website including the object detection results used in the annotation process (ii) to two expert workers.
Similar to the annotation process, we asked the expert workers to select bounding boxes from the auto-detected ones and modify them if they are not fitted to the tracked objects following one of the annotated referring expressions.
In Table~\ref{table:human_performance}, we present the human performance compared with the best model prediction, confirming the great performance gap between them.
The all image frames metrics of mSTIoU, mIoU+n and mAP@50+n have larger margins to the human performance than the metrics for REC of mIoU and mAP@50.
This suggests that the video clip-wise object localization from the referred expression is a much more difficult task than the simple REC task in 2D images.

\begin{table*}[t]
\begin{center}
    \footnotesize\begin{tabular}{lccccc}
    \toprule
    & \multicolumn{5}{c}{RefEgo Test} \\
    \cmidrule(lr){2-6}
    & \multicolumn{3}{c}{All images} & \multicolumn{2}{c}{Images w/ targets} \\
     \cmidrule(lr){2-4}\cmidrule(lr){5-6}
Model                & mSTIoU & mIoU+n & mAP@50+n & mIoU & mAP@50 \\
\midrule
OFA$^{\dagger}$     & 15.4 & 28.9 & 29.3 & 27.8 & 27.1 \\
OFA$^{\ddagger}$    & 31.7 & 44.4 &  47.5 & \textbf{51.0} & \textbf{56.2} \\
\midrule
MDETR+BH$^{\ddagger}$ (all)& 36.9 & \textbf{45.7} & \textbf{51.1} & 45.7 & 53.0 \\
~~~~+Object tracking & \textbf{37.6} & 45.4 & 51.0 & 46.0 & 53.4 \\
\midrule
Human                & 77.1 & 82.9 & 85.4 & 78.8 & 82.1 \\
~~~~$\Delta$         & +39.5& +37.2& +34.3& +27.8 & +25.9 \\
\bottomrule
	\end{tabular}
    \caption{
        Comparisons with human performance. $\Delta$ represents the difference between the human performance and the best model prediction (\textbf{bold}).
    }
    \label{table:human_performance}
\end{center}
\end{table*}

\end{document}